\documentclass[12pt]{article}
\author{Longfei Lu\footnote{Longfei.Lu@live.com}}
\usepackage{graphicx}
\usepackage{caption}
\usepackage{paralist} 
\usepackage[normalem]{ulem} 
\usepackage[urlcolor=blue,colorlinks=true]{hyperref}
\usepackage{subfigure}
\usepackage{float}
\usepackage{bm}
\usepackage{epsfig}
\usepackage[utf8]{inputenc}  

\newcommand{\ba}{\begin{equation}}
\newcommand{\ea}{\end{equation}}

\title{Optimal $\gamma$ and $C$ for $\epsilon$-Support Vector Regression with RBF Kernels}
\begin{document}
\bibliographystyle{plain}
\maketitle
\clearpage

\section*{Abstract}
The objective of this study is to investigate the efficient determination of $C$ and $\gamma$ for Support Vector Regression with RBF or mahalanobis kernel based on numerical and statistician  considerations, which indicates the connection between $C$ and kernels and demonstrates that the deviation of geometric distance of neighbour observation in mapped space effects the predict accuracy of $\epsilon$-SVR. We determinate the arrange of $\gamma$ \& $C$ and propose our method to choose their best values.

\section*{Introduction}
Traditional forecasting algorithm like ARIMA, Exponential Smoothing can provide good forecasting results with regard to trend, season and other linear correlated features . In practice, those features are normally non-linear. To solve non-linear forecasting problems, $\epsilon$-Support vector regression ($\epsilon$-SVR) is employed. Support vector machine (SVM) is nowadays wildly used for classification problems in many areas. However, $\epsilon$-SVR is hardly used because of the uncertain parameter $C$, $\epsilon$ for its dual problem. By using RBF or Mahalanobis kernels, value of $\gamma$ decides determination of kernel matrix and hence is the key of whole system.  An overview of choosing those parameters is given by \cite{L.Nr.64}. Best selection of $C$ is given by \cite{L.Nr.164}, which can be done with in limited iterations. The most used methods are searching methods like random search, grid search, pattern search. Cherkassy and Ma (2004)\cite{L.Nr.65} have proposed one way to determinate those parameters directly from the data. But there is still one parameter $c$ need to be extra searched within pre-existing arrange. Our propose is also to determinate $C$ and $\gamma$ directly but without using any searching methods. In this paper,
we give a short overview of $\epsilon$-SVR in section 1 and discuss the determination of $C$ and $\gamma$ in section 2 and 3. Then we test our algorithms using practice data in section 4 and summery results in section 5.

\section{$\epsilon$-Support Vector Regression}
Assume  $(x_{1},y_{1}),(x_{2},y_{2}),...,(x_{N},y_{N}),\dots\in \mathcal{X}\times\mathbf{R}\subseteq \mathbf{R}^{n+1}$ are observation pairs, $x_{i}\in\mathcal{X}\subset\mathbf{R}^n$ is feature vector and $y_{i}\in\mathbf{R}$ is the target output. Define $N_{\mathcal{X}}$ as the total number of all observation in training set. According to \cite{L.Nr.97} the dual problem form of $\epsilon$-SVR under given $C$ with kernel $K$ is
\begin{equation}
\label {FOR_OPTIMISATION_10}
\begin{array}{rc}
\mbox{\textbf{min}} & \displaystyle\frac{1}{2}(\underline{\alpha}-\underline{\alpha}^\ast)^{T}Q(\underline{\alpha}-\underline{\alpha}^\ast)
+\epsilon\sum_{}^{}(\alpha_{i}+\alpha_{i}^\ast) + \sum_{}^{}y_{i}(\alpha_{i}-\alpha_{i}^\ast)\\
\mbox{\textbf{subject to}} & \underline{e}^{T}(\underline{\alpha}-\underline{\alpha}^\ast) = 0,\\
  & 0\leq \alpha_{i},\alpha_{i}^{\ast}\leq C,i=1,...,N_{s},
\end{array}
\end{equation}
where $Q_{i,j}:= K(x_{i},x_{j})$ and $\underline{\alpha}=(\alpha_{1},...,\alpha_{N_{s}})$, $\underline{\alpha}^{\ast}=(\alpha_{1}^{\ast},...,\alpha_{N_{s}}^{\ast})$.
The corresponding approximate function is
\ba \label {FOR_OPTIMAL_00}
f(x) \;= \; \sum_{i=1}^{N_{s}}(-\alpha_{i}+\alpha_{i}^{*})K(x_{i},x)+b,
\ea
where $N_{s}$ is the total number of Support Vectors which are obtained from input parameter $\epsilon$. We consider $\epsilon$ as accuracy indicator or acceptable tolerance within this system. In practice, we use natural defined tolerance as input value for $\epsilon$.
\section{Optimal $\gamma$}

\begin{figure}[H]
	\centering
  \includegraphics[width=\textwidth, height=260px ]{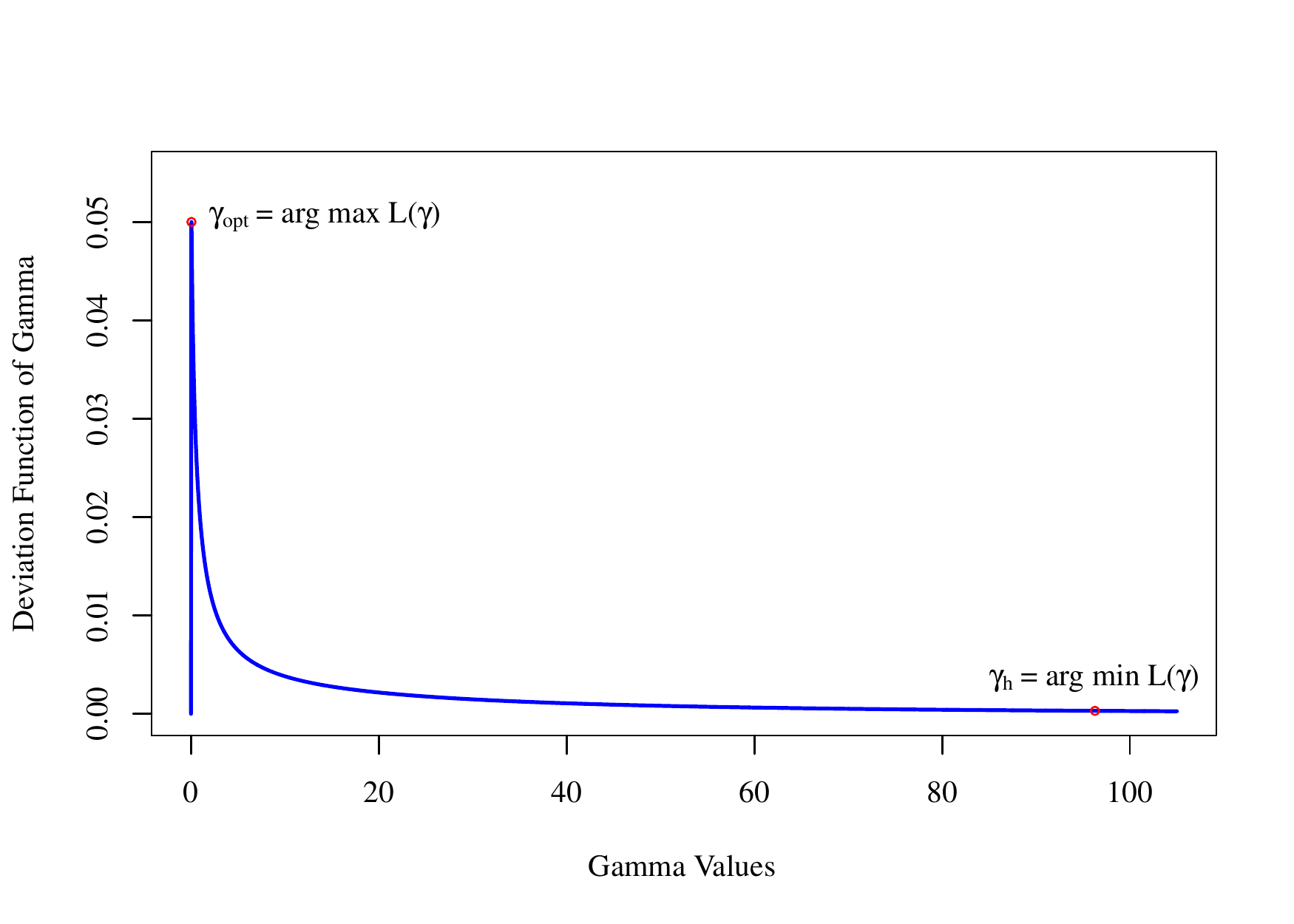}
	\caption{Deviation function $L(\gamma)$ for RBF kernel}
	\label{DeviFunGamma}
\end{figure}

\begin{figure}[H] 
\centering
\subfigure[$\gamma = \mbox{arg}\max L(\gamma)$] {\includegraphics[height=220px,width=0.85\textwidth]{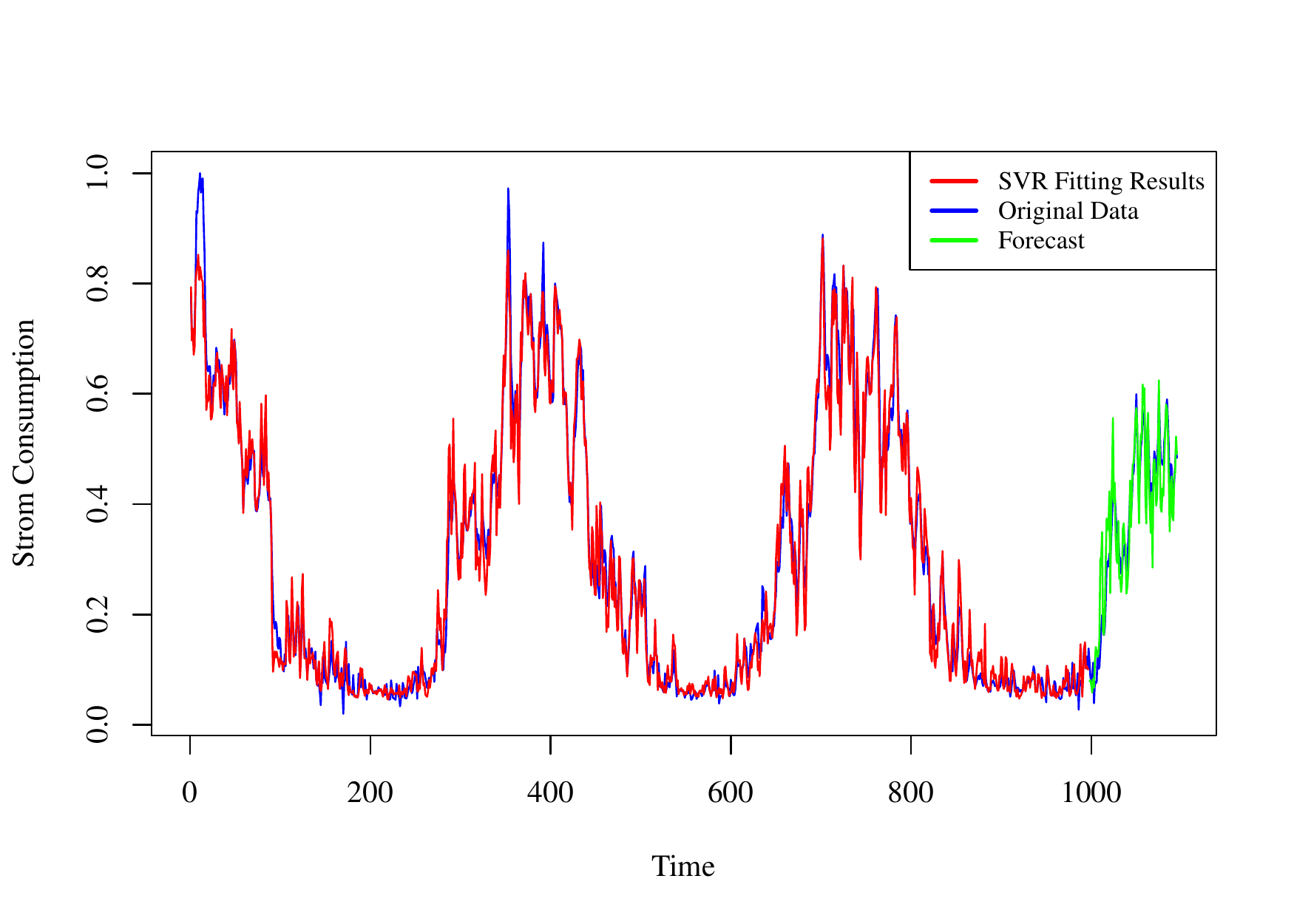}}
\subfigure[Using searching method] {\includegraphics[height=110px,width=0.45\textwidth]{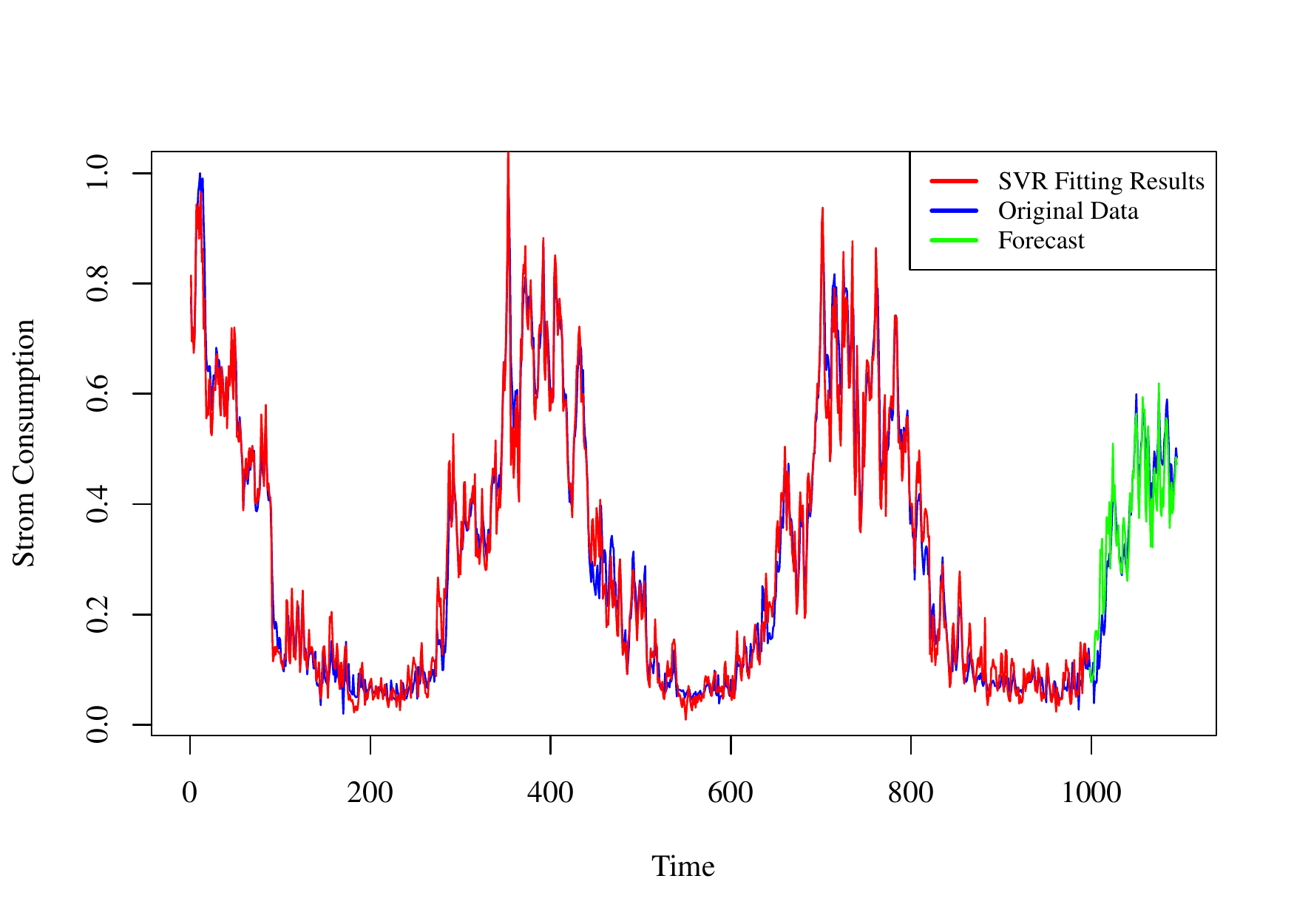}}
\subfigure[$\gamma = \mbox{arg}\min L(\gamma)$ (Over-fitting Problem)] {\includegraphics[height=110px,width=0.45\textwidth]{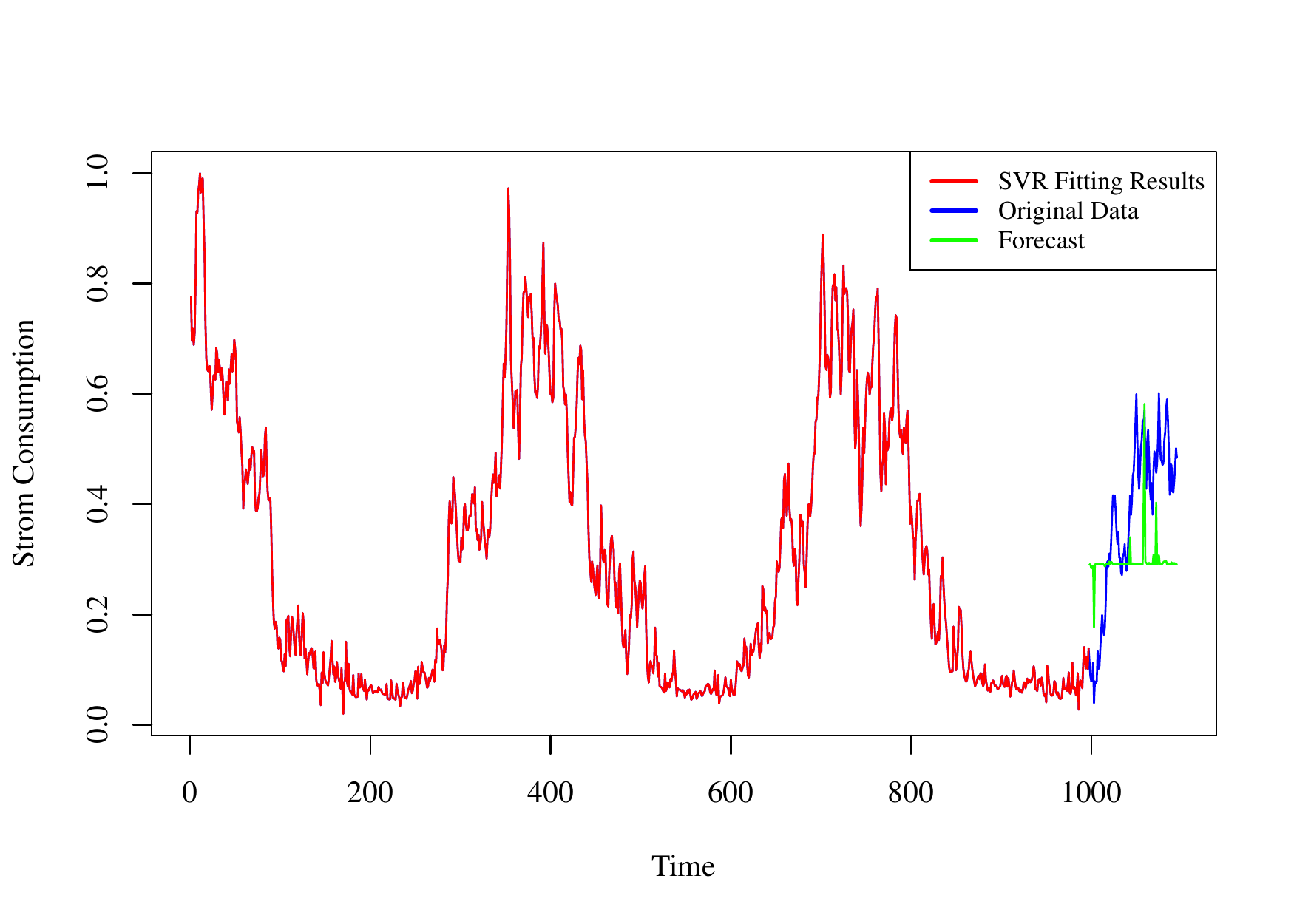}}
  \caption{ Results generated by package "e1071" in R}
  \label{Expriment}
\end{figure}

A feature map $\phi\colon 
x\mapsto\phi(x)\in\mathbf{R}^m,m\geq n$ builds new norm with respect to kernel $K$:
\begin{eqnarray*}
\Vert\phi(x_{i})-\phi(x_{j})\Vert ^2 
& = & K(x_{i},x_{i}) + K(x_{j},x_{j}) - 2K(x_{i},x_{j})\\
& = &  2 - 2K(x_{i},x_{j}),\mbox{ for RBF kernel}\\
& & \forall  x_{i},x_{j} \in \widehat{{X}}\subset \mathbf{R}^n.
\end{eqnarray*}
In mapped space, the splitting of two neighbour observation   , which means the deviation of mapped features should be large. Small values of deviation mean mapped features have very few difference and lead to linear dependences of kernel matrix in practice which produce few independent features and enlarge the solutions space. it is not difficult to show that features in RBF Kernel mapped features have same structure as in original which means $$\Vert x_{i} - x_{j} \Vert \leq \Vert x_{k} - x_{l} \Vert  \Leftrightarrow  \Vert \phi(x_{i}) - \phi(x_{j}) \Vert \leq \Vert \phi(x_{k}) - \phi(x_{l}) \Vert.$$  More independence of mapped features lead to small solution spaces. One measure for this character is the deviation of all 2-te Norms between every two observation. Define \textbf{Deviation Function} of observation as
$$ \bm{L(\gamma)} :=
\frac{1}{N_{\mathcal{X}}(N_{\mathcal{X}}-1)}\sum_{(i,j)\in\mathcal{X}}^{}\bigg(\Vert \phi(x_{i})-\phi(x_{j})\Vert - \frac{1}{N_{\mathcal{X}}(N_{\mathcal{X}}-1)}\sum_{(k,l)\in\mathcal{X}}^{}\Vert \phi(x_{k})-\phi(x_{l}) \Vert \bigg)^2.
$$
The Deviation Function ${L(\gamma)}$ describes geographic differences of whole observation in mapped space with help of 2-te Norm and is convex. Our best choice of $\gamma_{opt}$ is the point when the $L(\gamma)$ reaches its maximum where $L^{'}(\gamma_{opt})=0$ with fist derivation formal
\begin{eqnarray*}
L^{'}(\gamma) & = & \frac{2}{N_{\mathcal{X}}(N_{\mathcal{X}}-1)} \sum_{(i,j)\in\mathcal{X}}^{}\bigg( \widehat{L}(x_{i},x_{j})-\frac{1}{N_{\mathcal{X}}(N_{\mathcal{X}}-1)}\sum_{(k,l)}^{}\widehat{L}(x_{k},x_{l}) \bigg) \\
& &\cdotp \bigg( \widehat{D}(x_{i},x_{j})-\frac{1}{N_{\mathcal{X}}(N_{\mathcal{X}}-1)}\sum_{(k,l)\in\mathcal{X}}^{}\widehat{D}(x_{k},x_{l}) \bigg),
\end{eqnarray*}
and its second derivation formal is
\begin{eqnarray*}
L''(\gamma)&=&\frac{2}{N_{\mathcal{X}}(N_{\mathcal{X}}-1)}\sum\limits_{(i,j)\in\mathcal{X}}\left(\widehat{L}'(x_i,x_{j})-\frac{1}{N_{\mathcal{X}}(N_{\mathcal{X}}-1)}\sum\limits_{(k,l)\in\mathcal{X}}\widehat{L}'(x_{k},x_{l})\right) \\
&& \cdot \left(\widehat{D}(x_i,x_{j})-\frac{1}{N_{\mathcal{X}}(N_{\mathcal{X}}-1)}\sum\limits_{(k,l)\in\mathcal{X}}\widehat{D}(x_{k},x_{l})\right) \\
&+&\frac{2}{N_{\mathcal{X}}(N_{\mathcal{X}}-1)}\sum\limits_{(i,j)\in\mathcal{X}}\left(\widehat{L}(x_i,x_{j})-\frac{1}{N_{\mathcal{X}}(N_{\mathcal{X}}-1)}\sum\limits_{(k,l)\in\mathcal{X}}\widehat{L}(x_{k},x_{l})\right) \\
&& \cdot \left(\widehat{D}'(x_i,x_{j})-\frac{1}{N_{\mathcal{X}}(N_{\mathcal{X}}-1)}\sum\limits_{(k,l)\in\mathcal{X}}\widehat{D}'(x_{k},x_{l})\right) \\
&=&\frac{2}{N_{\mathcal{X}}(N_{\mathcal{X}}-1)}\sum\limits_{(i,j)\in\mathcal{X}}\left(\widehat{D}(x_i,x_{j})-\frac{1}{N_{\mathcal{X}}(N_{\mathcal{X}}-1)}\sum\limits_{(k,l)\in\mathcal{X}}\widehat{D}(x_{k},x_{l})\right)^2\\
&+&\frac{2}{N_{\mathcal{X}}(N_{\mathcal{X}}-1)}\sum\limits_{(i,j)\in\mathcal{X}}\left(\widehat{L}(x_i,x_{j})-\frac{1}{N_{\mathcal{X}}(N_{\mathcal{X}}-1)}\sum\limits_{(k,l)\in\mathcal{X}}\widehat{L}(x_{k},x_{l})\right) \\
&& \cdot \left(\widehat{D}'(x_i,x_{j})-\frac{1}{N_{\mathcal{X}}(N_{\mathcal{X}}-1)}\sum\limits_{(k,l)\in\mathcal{X}}\widehat{D}'(x_{k},x_{l})\right),
\end{eqnarray*}
with respect to
\begin{eqnarray*}
\widehat{L}(x_{i},x_{j})& = & \bigg( 2 - 2 \exp(-\gamma G(x_{i},x_{j})) \bigg)^{\frac{1}{2}} \\
 \widehat{D}(x_{i},x_{j})& = & G(x_{i},x_{j})   \exp(-\gamma G(x_{i},x_{j}))\big( 2 - 2\cdotp\exp(-\gamma G(x_{i},x_{j})\big)^{-\frac{1}{2}}.
\end{eqnarray*}

\begin{eqnarray*}
\widehat{D}'(x_i,x_{j})&=&-(G(x_i,x_{j}))^2\exp(-\gamma G(x_i,x_{j}))\left(2-2\exp(-\gamma G(x_i,x_{j}))\right)^{-\frac{1}{2}}\\
&&\cdot \left(1+\exp(-\gamma G(x_i,x_{j}))(2-2\exp(-\gamma G(x_i,x_{j})))^{-1}\right)\\
&=&-G(x_i,x_{j})\widehat{D}(x_i,x_{j})\left(1+\exp(-\gamma G(x_i,x_{j}))(\widehat{L}(x_i,x_{j}))^{-2}\right)\\
\end{eqnarray*}
For RBF Kernel $K(x_{i},x)=\exp(-\gamma \Vert x_{i} - x  \Vert^2)$: 
\begin{eqnarray*}
 G(x_{i},x_{j}) & = & \Vert x_{i} - x_{j}\Vert^{2}. 
 \end{eqnarray*}
For Mahalanobis Kernel $K(x_{i},x)=\exp(-\gamma \frac{1}{m}(x_{i}-x)Q^{-1}(x_{i}-x))$: 
 \begin{eqnarray*}  
   G(x_{i},x_{j}) & = & \frac{1}{m}(x_{i}-x_{i+1})Q^{-1}(x_{i}-x_{j}),
 \end{eqnarray*}
with
 \begin{eqnarray*}
   Q & = & \frac{1}{N_{\mathcal{X}}}\sum_{k=1}^{N_{\mathcal{X}}}(x_{k}-c)(x_{k}-c)^{T},\\
   c & = & \frac{1}{N_{\mathcal{X}}}\sum_{k=1}^{N_{\mathcal{X}}}x_{k},\\
   m & = & \frac{1}{N_{\mathcal{X}}}\sum_{k=1}^{N_{\mathcal{X}}}(x_{k}-c)^{T}Q^{-1}(x_{k}-c).
\end{eqnarray*}
Because of the convexity of deviation function, its maximum always exists and by using numeric method like Newton Method we obtain very good balance between computation performance and model accuracy. Our proposed deviation function $L(\gamma)$ is also very helpful for determining arrange for searching method (see figure \ref{DeviFunGamma}). The arrange of $\gamma$ is $(0,\gamma_{h})$, where $\gamma_{h}=\mbox{arg min}L(\gamma)$. At point $\gamma_{h}$, the value of kernel function $K(x_{i},x_{j})\approx 0, \mbox{ for } ~i\neq j$, which leads to strong over-fitting problem (see figure \ref{Expriment}). The best value of $\gamma$ is at $\mbox{arg max } L(\gamma)$.

\section{Optimal choice of \boldmath $C$}

We use the proposed method from  \cite{L.Nr.164} to get best value of $C$, which provides very stable solution. In order to reduce the total number of iterations, we give a reasonable initial $C$ before it begins by applying mean value theorem in mapped space. From (\ref{FOR_OPTIMAL_00}) exists $x\in\mathcal{X}$ satisfying 
$$\Vert y_{i}-y_{j}\Vert = \Vert f'(\phi(x))\cdot(\phi(x_{i})-\phi(x_{j}))\Vert,\forall i,j\in\{1,..,N_{\mathcal{X}}\}.$$
For RBF Kernel, we can proof that
\ba \label {ProofOne}
\frac{\Vert  y_{i}-y_{j} \Vert}{\Vert \phi(x_{i})- \phi(x_{j}) \Vert}\leq \Vert f'(\phi(x)) \Vert \leq N_{s} C.
\ea
For worst situation that $N_{s}=1$ and $\Vert \phi(x_{i})- \phi(x_{j}) \Vert= \exp(-\gamma\Vert x_{i} - x_{j} \Vert ^2 )$, we obtain
\ba  \label {InitialC}
C \geq \max_{i,j\in\{1,..,N_{\mathcal{X}}\}}\Vert  y_{i}-y_{j} \Vert\exp(\gamma\Vert x_{i} - x_{j} \Vert ^2 ).
\ea 
We use right side of (\ref{InitialC}) as initial $C_{ini}$ for iteration of finding best $C$. New value of $C$ after every solving of SVR is calculated by 
\ba \label {NewC}
	C_{new} \; = \; \frac{N_{\mathcal{X}}+N_{s}}{\sum\limits_{x_{i}\in X_{C}}L_{\epsilon}(y_{i}-f(x_i))+\sum\limits_{x_i\in X_{M}}\frac{1}{C- \vert\alpha_{i}-\alpha_{i}^{*}\vert} + \frac{\epsilon N_{\mathcal{X}}}{\epsilon C+1}}
\ea
with
\begin{eqnarray*}
X_{C}& = &\{x_i|\;\vert y_i-f(x_i)\vert > \epsilon \mbox{ with }\alpha_i=C,\alpha^{*}_i=0\mbox{ or }\alpha_i=0,\alpha^{*}_i=C\}, \\
X_{M}& = &\{x_i|\;\vert y_i-f(x_i)\vert = \epsilon \mbox{ with } 0<\alpha_i< C ,\mbox{ or } 0 <\alpha^{*}_i < C\}
\end{eqnarray*}
and $\epsilon$ loss function $L_{\epsilon}$ is defined as
\ba \label {EpsilonLossFunction}
 L_{\epsilon}(y_i-f(x_i)) =\left\{\begin{array}{cl} 0 & \mbox{ for }\vert y_i - f(x_i)\vert \leq \epsilon \\
\vert y_i - f(x_i) \vert - \epsilon & \mbox{ otherwise.}\end{array}\right. 
\ea

\begin{figure}[H]
	\centering
 \includegraphics[height=240px,width=0.95\textwidth]{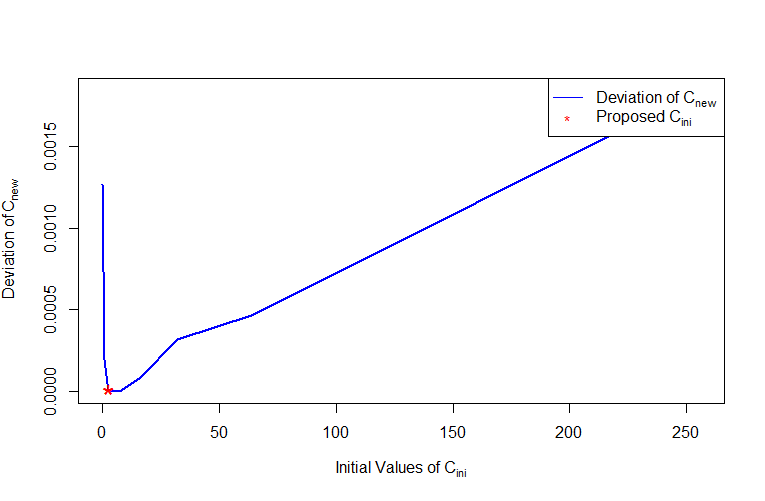}
	 \caption{Devation of $C_{new}$ with respect to different initial value $C_{ini}$}
	 	\label{C_new}
\end{figure}

\begin{figure}[H]
	\centering
 \includegraphics[height=240px,width=0.95\textwidth]{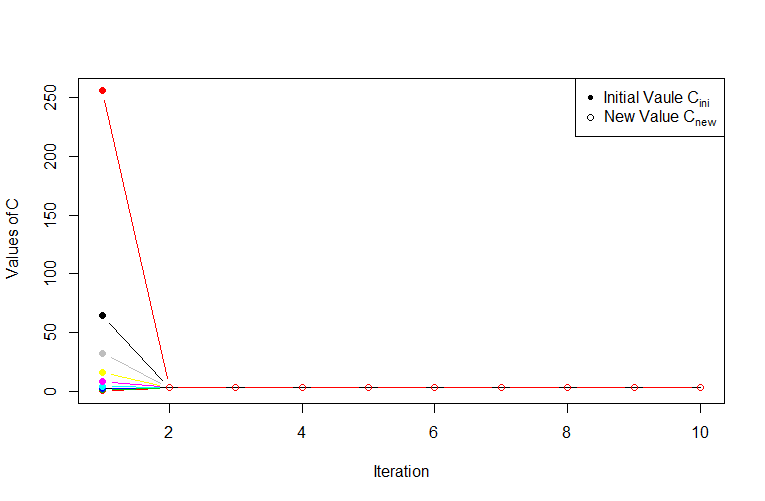}
	 \caption{Iteration of finding stable $C$ with different $C_{ini}$}
	 	\label{IterationOfC}
\end{figure}

According to our experiments, our proposed initial value has relative small deviation during computation of $C_{new}$ (see figure \ref{C_new} ), which reduces the iteration number by setting stop critical as $\Delta$ changes between new and last $C_{new}$. Experiments also show that no matter how big $C_{ini}$ is, it archives its stable value within limit iterations. Figure \ref{IterationOfC} and Table \ref{IterationTable} show that $C$ value changes not so much just after second solving SVR in our experiments by input data of gas consumption with temperature and weekdays as features. 

\begin{table}[H]
\begin{center}
\scalebox{0.55}{
\begin{tabular}{|c|c|c|c|c|c|c|c|c|c|c|} \hline
\multicolumn{1}{|c|}{$C_{ini}$}&\multicolumn{1}{c|}{1st $C_{new}$}&\multicolumn{1}{c|}{2nd $C_{new}$}&\multicolumn{1}{c|}{3rd $C_{new}$}&\multicolumn{1}{c|}{4th $C_{new}$}&\multicolumn{1}{c|}{5th $C_{new}$}&\multicolumn{1}{c|}{6th $C_{new}$}&\multicolumn{1}{c|}{7th $C_{new}$}&\multicolumn{1}{c|}{8th $C_{new}$}&\multicolumn{1}{c|}{9th $C_{new}$}&\multicolumn{1}{c|}{10th $C_{new}$}\\ \hline
~~2.582123&~~2.887401&~~2.895912&~~2.893142&~~2.903892&~~2.898490&~~2.898554&~~2.898552&~~2.898462&~~2.898543&~~2.893128\\ 
~~0.100000&~~2.738660&~~2.897665&~~2.898563&~~2.898521&~~2.898552&~~2.898515&~~2.898516&~~2.898517&~~2.898550&~~2.893649\\ 
~~1.000000&~~2.836104&~~2.893117&~~2.893175&~~2.903887&~~2.898516&~~2.898489&~~2.898544&~~2.898492&~~2.898560&~~2.898525\\ 
~~2.000000&~~2.866985&~~2.898480&~~2.898493&~~2.898526&~~2.898519&~~2.898527&~~2.898528&~~2.898566&~~2.898478&~~2.898450\\ 
~~4.000000&~~2.891488&~~2.898533&~~2.898536&~~2.898512&~~2.898500&~~2.898469&~~2.898476&~~2.898543&~~2.898502&~~2.898529\\ 
~~8.000000&~~2.892421&~~2.893184&~~2.898528&~~2.898532&~~2.898523&~~2.898495&~~2.898494&~~2.898525&~~2.898560&~~2.898519\\ 
~16.000000&~~2.936898&~~2.893208&~~2.893217&~~2.893139&~~2.903910&~~2.898495&~~2.898543&~~2.898531&~~2.898510&~~2.898463\\ 
~32.000000&~~2.977581&~~2.897040&~~2.898510&~~2.898527&~~2.898548&~~2.897689&~~2.898522&~~2.898533&~~2.898511&~~2.898530\\ 
~64.000000&~~2.995011&~~2.905414&~~2.902729&~~2.898555&~~2.898485&~~2.898533&~~2.898483&~~2.898527&~~2.898541&~~2.898588\\ 
256.000000&~~3.090617&~~2.909661&~~2.898573&~~2.898489&~~2.898521&~~2.898504&~~2.898506&~~2.898570&~~2.898555&~~2.898496\\ 
\hline
\end{tabular}}
\caption{Iterations Table}
\label{IterationTable}
\end{center}
\end{table}

\section{Experiments}

All experiments were performed by using practice data from an energy company from 2009-01-01 to 2011-12-31, which contains 15 different features. The training set was set from 2009-01-01 to 2011-09-24. We determinated the curves of $\gamma$ according to the description in section 3 and used $\gamma$ at maximum of $L(\gamma)$ for determination of parameter $C$. Back test was performed by using data from 2011-09-25 to 2011-12-31 for RBF kernels. Those results were compared with results generated by searching $\gamma \in\{2^{-15},...,2^{3}\}$ and $C\in\{2^{-5},...,2^{15}\}$. Package "e1071" of R was used. We scaled arrange of data into $[0,1]$ and used default scale option for further calculation. We employed root-mean-square error (RMSE) 
$$\mbox{RMSE}=\sqrt{\frac{\sum_{i=1}^{N}(y_{i}-f(x_{i}))^2}{N}}$$
and mean absolute percentage error (MAPE)
$$\mbox{MAPE}=\frac{100\%}{N}\sum_{i=1}^{N}\Big\vert\frac{y_{i}-f(x_{i})}{y_{i}}\Big\vert.$$
 as measures of accuracy of $\epsilon$-SVR. Except those two famous measures, we define a total new measure \textbf{System Error Deviation Measure} (SEDM) as
 \begin{equation}
 	\mbox{SEDM} = \frac{1}{N}\sum_{i=1}^{N}\bigg(\big(y_{i}-f(x_{i})\big)-\frac{1}{N}\sum_{j=1}^{N}\big(y_{j}-f(x_{j})\big)\bigg).
 \end{equation}

\begin{table}[H]
\centering
\scalebox{0.89}{
\begin{tabular}{lllllll}
\hline
& & &\multicolumn{2}{ c }{\textbf{Training}} &\multicolumn{2}{ c }{\textbf{Back Test}}  \\ \cline{4-5} \cline{5-7}
 	& $\gamma$ & $C$ & RMSE & MAPE & RMSE & MAPE\\ \cline{2-7}
 Our Propose: & 0.03224 & 0.42112 & 0.03808 &13.5430\% & 0.05082 &15.4615\% \\
 Tune Method: & 0.04000& 0.99000 & 0.03664 & 12.6075\%& 0.05262 & 15.7146\% \\
 Best Solution: &0.02 & 0.61 & 0.03758&13.6883\% &0.05005 & 14.9604\%\\
 \hline
  Over-fitting: &96.27366&55.30339 & 5.06542e-05&0.0279\% &0.16210 &55.3389\% \\
 \hline
\end{tabular}}
\caption{Results with $\epsilon=0$ for figure \ref{SolutionSpace_Gamma}}
\label{Results}
\end{table}

\begin{figure}[H]
	\centering
 \includegraphics[height=240px,width=0.95\textwidth]{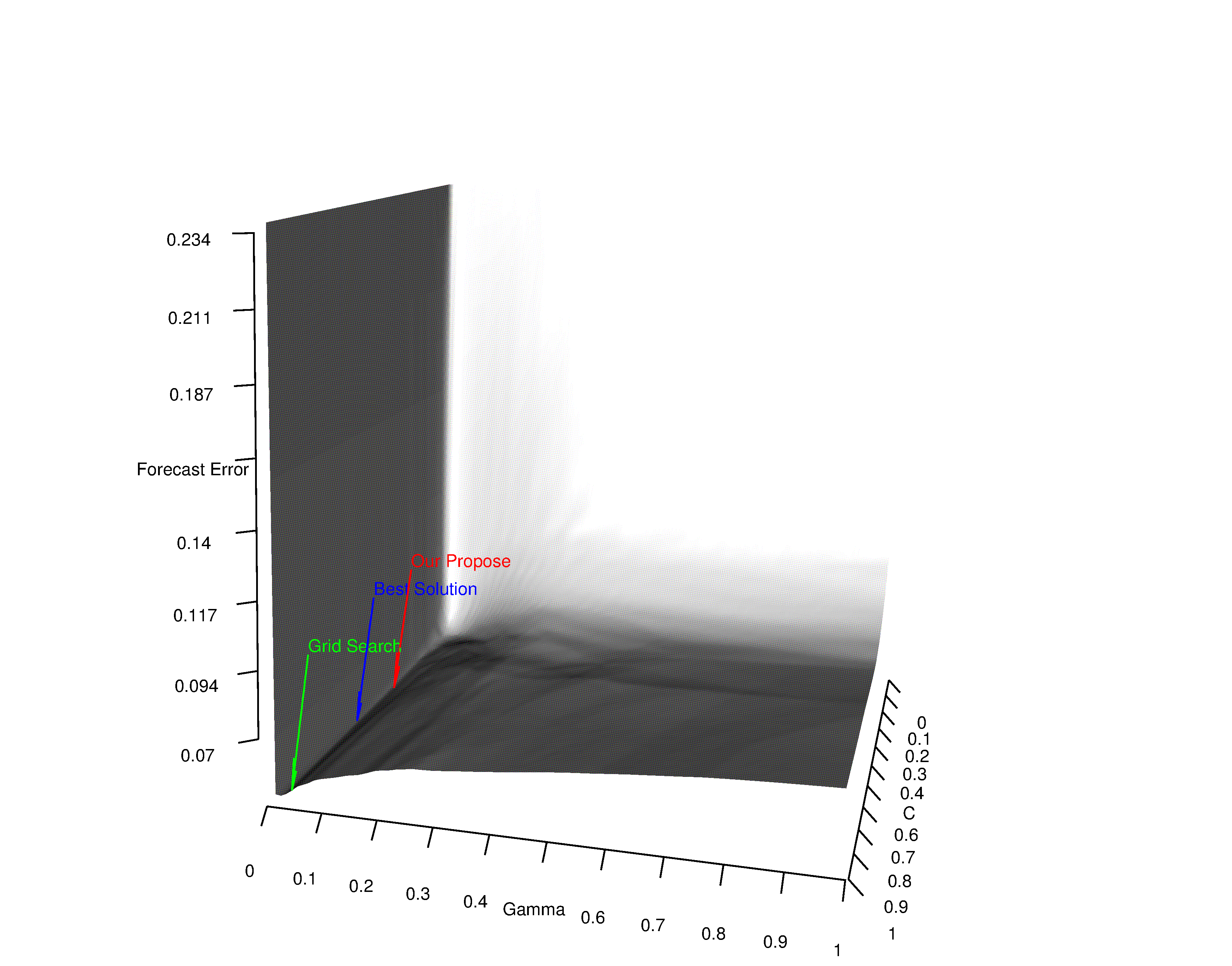}
	 \caption{Solution Space $\epsilon = 0$ for Table \ref{Results}}
	 	\label{SolutionSpace_Gamma}
\end{figure}

Table \ref{Results} shows that our propose produced better results than tune method, which has solved SVR over 10,000 times and has the smallest value in training area. The $\gamma$ value of best solution is smaller than our propose, which always has the smallest RMSE value in back test area. Setting of over-fitting is $\gamma=\mbox{arg min } L(\gamma)$. Figure \ref{SolutionSpace_Gamma} shows the whole $\gamma$-$C$-RMSE space. Our propose locates in the same level as grid search and best solution.

Table \ref{ResultsEps} shows the case that $\epsilon\neq 0$. We tuned  all three parameter extra within $[10^{-10},1]\times[10^{-10},1]\times[0,1]$ for $\gamma,C,\mbox{ and }\epsilon$, which has solved SVR over 1,000,000 times. We chose smallest RMSE value in training for Extra Tuning. But its forecasting error is worse than results in Table \ref{Results}. Therefore, we used the $\gamma_{best}$ from Best Solution, $\gamma_{opt}$ from Our Propose in Table \ref{Results} and each was tuned within interval $[10^{-10},1]$ for $C$ and $[0,0.99]$ for $\epsilon$. Our propose for $\gamma_{opt}$ produced better results in Table \ref{ResultsEps} with $\epsilon = 0.00516$ than Tune Method, which produced better results for $\gamma_{best}$. Figure \ref{SolutionSpace_Gamma} shows locations of our experimented solutions in $C$-$\epsilon$-RMSE space for each $\gamma_{opt}$ and $\gamma_{best}$.

\begin{table}[H]
\centering
\scalebox{0.85}{
\begin{tabular}{llllllll}
\hline
& & & &\multicolumn{2}{ c }{\textbf{Training}} &\multicolumn{2}{ c }{\textbf{Back Test}}  \\ \cline{5-6} \cline{6-8}
 	& $\gamma$ & $C$ & $\epsilon$ & RMSE & MAPE & RMSE & MAPE\\ \cline{2-8}
 Our Propose\\
 $\gamma_{opt}$ & 0.03224 & 0.42112 & 0.00516 & 0.03808 & 13.5563\% & 0.05079 &15.4560\%\\
 $\gamma_{best}$ & 0.02 & 0.22035 & 0.02768 & 0.03967 & 14.9641\%& 0.05038 & 15.6418\% \\
 \hline
 Tune Method\\
 $\gamma_{opt}$ & 0.03224 & 0.99 & 0.07 & 0.03635 & 14.0366\%& 0.05185 & 15.9645\% \\
  $\gamma_{best}$ & 0.02& 0.99 & 0.05 & 0.03681 & 13.9567\%& 0.05013 & 14.7957\% \\
 \hline\\
 Best Solution \\
 $\gamma_{opt}$ &0.03224 & 0.16 & 1e-10 & 0.04058 &  14.4724\% & 0.05026 & 15.5602\% \\
 $\gamma_{best}$ & 0.02 &0.71 &0.05 &0.03725 &14.1742\% &0.04992 &14.8701\% \\ \hline
 
 Extra tuning &  0.06 & 0.99 & 0.09&0.03599 &14.3002\% &0.05584& 18.0309\%\\ \hline
\end{tabular}}
\caption{Results with optimal $\epsilon$ for figure \ref{SolutionSpace}}
\label{ResultsEps}
\end{table}

\begin{figure}[H]
	\centering
	\subfigure[$\gamma_{opt}$]
  {\includegraphics[height=140px,width=0.40\textwidth]{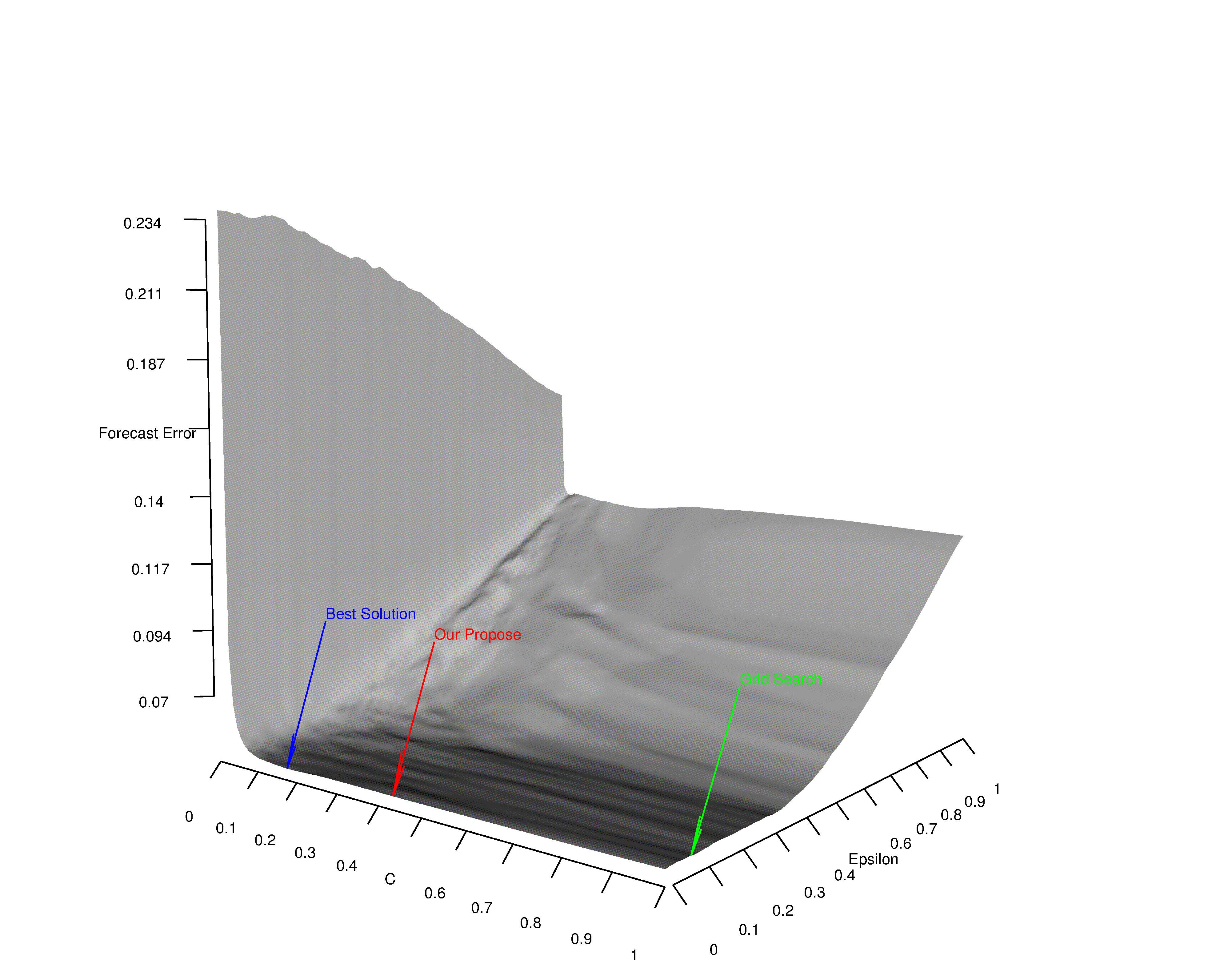}	}
 \subfigure[ $\gamma_{best}$ ]
 { \includegraphics[height=140px,width=0.45\textwidth]{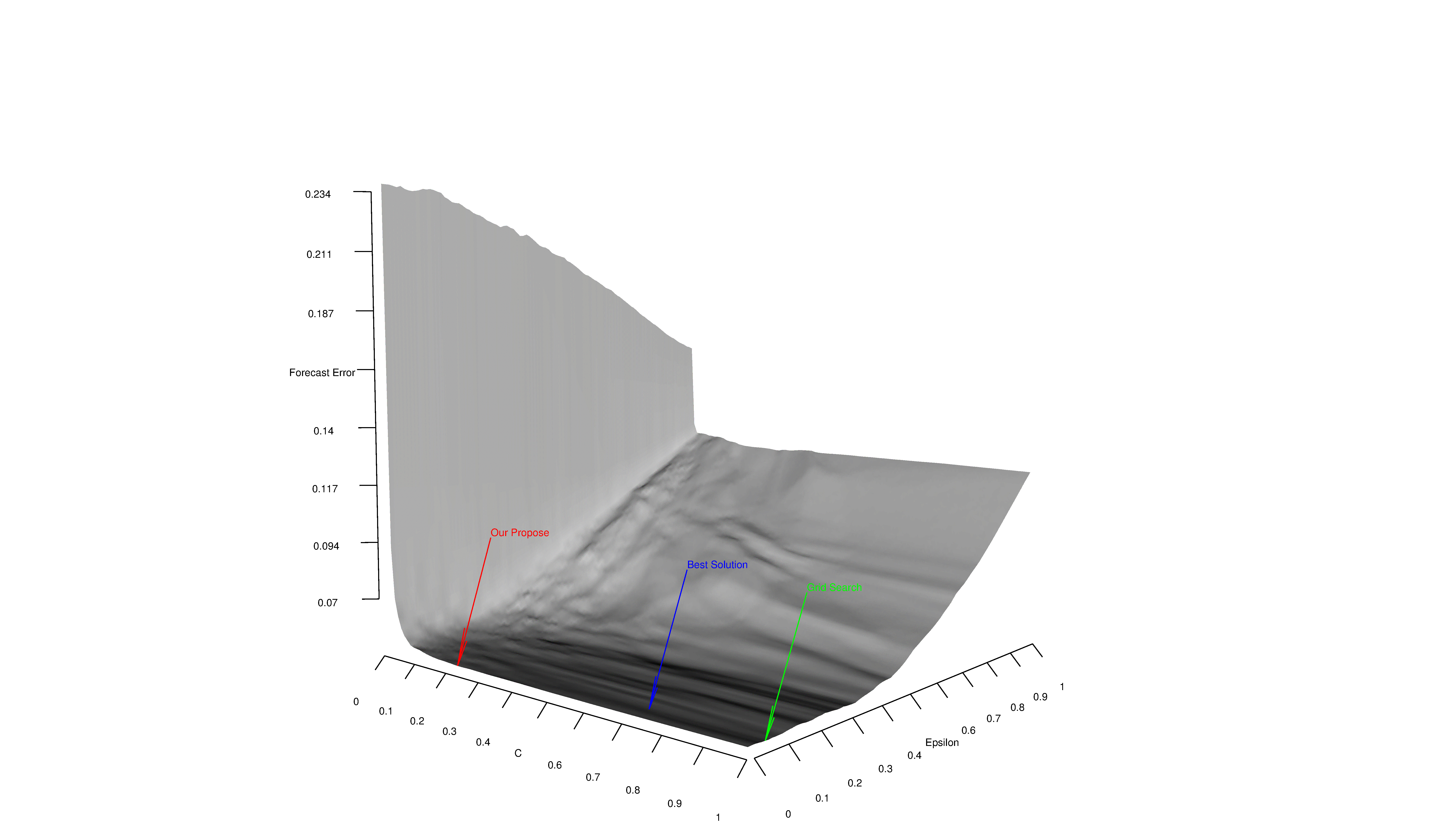}}
	 \caption{Solution Space for Table \ref{ResultsEps}}
	 	\label{SolutionSpace}
\end{figure}

\section{Conclusion}
Deviation function $\bm L(\gamma)$ and set $\bm C$ give the arrange of $\gamma$ and $C$, also provide possible solutions to choose their optimal values. By searching method, RMSE of training set became smaller than before. 

\bibliography{OptimalParam}
\end{document}